\documentclass[letterpaper]{article} 
\usepackage[draft]{aaai25}  
\usepackage{times}  
\usepackage{helvet}  
\usepackage{courier}  
\usepackage[hyphens]{url}  
\usepackage{graphicx} 
\usepackage{natbib}  
\usepackage{caption} 
\usepackage{amsmath, amsthm, amsfonts, mathtools}
\usepackage{booktabs}
\usepackage{subcaption}

\usepackage{algorithm}
\usepackage{algorithmic}

\usepackage{newfloat}
\usepackage{listings}
\DeclareCaptionStyle{ruled}{labelfont=normalfont,labelsep=colon,strut=off} 
\floatstyle{ruled}
\newfloat{listing}{tb}{lst}{}
\floatname{listing}{Listing}
\newtheorem{theorem}{Theorem}

\newcommand{\calib}{\textrm{cal}}
\newcommand{\test}{\textrm{test}}
\newcommand{\train}{\textrm{train}}
\newcommand{\res}{\textrm{res}}
\newcommand{\locart}{\textrm{locart}}

\title{PersonalizedUS: Interpretable Breast Cancer Risk Assessment with Local Coverage Uncertainty Quantification}
\author {
    Alek Fröhlich\textsuperscript{\rm 1}\equalcontrib,
    Thiago Ramos\textsuperscript{\rm 2}\equalcontrib,
    Gustavo Cabello\textsuperscript{\rm 3},
    Isabela Buzatto\textsuperscript{\rm 3},
    Rafael Izbicki\textsuperscript{\rm 2},
    Daniel Tiezzi\textsuperscript{\rm 3}
}
\affiliations {
    \textsuperscript{\rm 1}UFSC, Florianópolis, Brazil\\
    \textsuperscript{\rm 2}UFSCar, São Carlos, Brazil\\
    \textsuperscript{\rm 3}USP, Ribeirão Preto, Brazil\\
    alek.frohlich@posgrad.ufsc.br,
    thiagorr@ufscar.br,
    gustavocabello@usp.br,
    ipcarlotti@hcrp.usp.br,
    rizbicki@ufscar.br, dtiezzi@usp.br
}

\begin{document}
\maketitle
\begin{abstract}
Correctly assessing the malignancy of breast lesions identified during ultrasound examinations is crucial for effective clinical decision-making. However, the current ``golden standard'' relies on manual BI-RADS scoring by clinicians, often leading to unnecessary biopsies and a significant mental health burden on patients and their families. In this paper, we introduce PersonalizedUS, an interpretable machine learning system that leverages recent advances in conformal prediction to provide precise and personalized risk estimates with local coverage guarantees and sensitivity, specificity, and predictive values above 0.9 across various threshold levels. In particular, we identify meaningful lesion subgroups where distribution-free, model-agnostic conditional coverage holds, with approximately 90\% of our prediction sets containing only the ground truth in most lesion subgroups, thus explicitly characterizing for which patients the model is most suitably applied. Moreover, we make available a curated tabular dataset of 1936 biopsied breast lesions from a recent observational multicenter study and benchmark the performance of several state-of-the-art learning algorithms. We also report a successful case study of the deployed system in the same multicenter context. Concrete clinical benefits include up to a 65\% reduction in requested biopsies among BI-RADS 4a and 4b lesions, with minimal to no missed cancer cases.
\end{abstract}

\section{Introduction}

Breast cancer is the most diagnosed cancer and the leading cause of cancer-related deaths among women worldwide, with 2.3 million new cases and 666,000 deaths reported in 2022 \cite{Bray2024}. While screening mammography is widely used to increase early diagnosis—when treatment is more effective and less costly \cite{Prager2018}—it has limitations, particularly a high false positive rate, leading to recalls, short interval follow-ups, and unnecessary biopsies \cite{Hubbard2011}. A recent study estimated that, after 10 years of annual screening in women aged 40-59 years, 61\% of individuals would experience at least one false positive recall and up to 9\% at least one false positive breast biopsy \cite{Ho2022}.

Breast ultrasound (US) is broadly used as a diagnostic tool complementing inconclusive mammograms, evaluating palpable findings, and guiding breast biopsies \cite{Berg2008}. Breast US has several advantages compared to other imaging modalities, including relatively lower cost, lack of ionizing radiation, and real-time evaluation capabilities \cite{Feig2010}.

Despite these advantages, interpreting US exams is hard. The evaluation involves inspecting image features such as lesion size, shape, and margins for signs of malignancy, and relies heavily on the operator's experience \cite{Sivarajah2020}. Radiologists ultimately decide whether the findings are benign, need follow-up, or require a biopsy according to the Breast Imaging Reporting and Data System (BI-RADS) \cite{BIRADS_US}. Although the BI-RADS standardized the description of lesions and reports, it has notable limitations. In particular, the most recent version offers little guidance on how to subdivide BI-RADS 4 findings into subcategories (a, b, and c), with risk ranging from 5\% to 95\%, and overlooks clinical and Doppler-based features known to be important for malignancy prediction \cite{Pfob2022}. These gaps contribute to the method's intra-reader variability and false positive rate \cite{Lazarus2006, Niu2020}. Notably, incorporating US in breast cancer screening leads to an additional 4-8\% of patients undergoing biopsy as compared to mammography alone, yet only 7-8\% of US-guided breast biopsies are found to be malignant \cite{Yang2020, Corsetti2011}.

The main goals of our work were to:
\begin{itemize}
    \item Improve upon the current BI-RADS standard by reducing the number of false positives without compromising cancer cases;
    \item Provide an interpretable model that can be safely monitored by medical doctors with suitable training in AI;
    \item Give personalized uncertainty quantification for model predictions with strong theoretical guarantees under minimal assumptions;
    \item Provide an intuitive interface for doctors to interact with the model.
\end{itemize}

\begin{figure*}[t]
    \centering
    \includegraphics[width=0.9\textwidth]{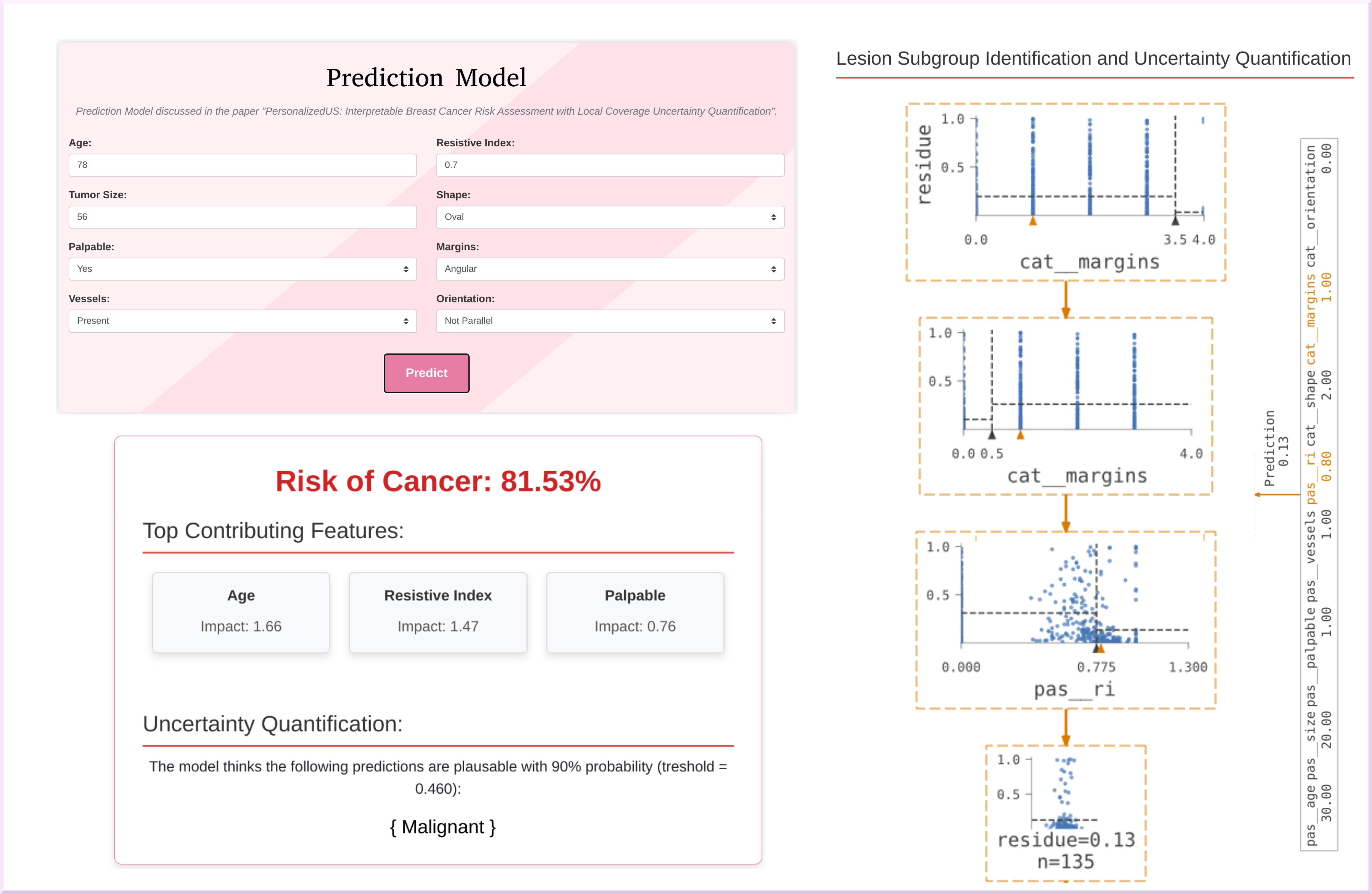}
    \caption{PersonalizedUS pipeline: Starting with a suspicious breast lesion identified by US, clinical, BI-RADS, and Doppler features are fed into a logistic regression model to estimate malignancy risk. The most influential features are highlighted. The lesion is then fed into a decision tree, whose leaves reflect difficulty levels with respect to the prediction model. Finally, a prediction set with a local conditional coverage guarantee, ensuring that lesions within more challenging leaves are associated with more uncertain predictions.}
    \label{fig:deployed_system}
\end{figure*}

To address these challenges, we present a solution that integrates traditional statistical methods with modern AI techniques, with a strong emphasis on uncertainty quantification. Concretely, we developed PersonalizedUS, an interpretable and accurate logistic regression model deployed in the form of an easy-to-use web application. The key innovation lies in the novel use of Locart, a recently developed conformal prediction method that leverages feature space partitions to provide personalized confidence intervals/sets that approach conditional coverage. This represents a significant advance in applying distribution-free, model-agnostic uncertainty quantification tools to the medical field, as Locart delivers localized uncertainty estimates that align more closely with the goals of precision medicine embodied by clinical prediction models.

\subsubsection{Contributions Overview}
\begin{itemize}
    \item \textbf{PersonalizedUS:}  We introduce PersonalizedUS, an interpretable machine learning system designed to accurately assess the malignancy risk of breast lesions detected by ultrasound imaging. By integrating a logistic regression model with modern conformal prediction techniques, PersonalizedUS offers precise and personalized risk estimates that improve upon the current BI-RADS standard, significantly reducing the number of false positives without missing cancer cases.

    \item \textbf{Novel Use of Conformal Prediction:} We  adapt the Locart method, a recent approach from conformal prediction that guarantees local conditional coverage. The method was originally developed for regression problems and is applied to binary classification for the first time here.
\end{itemize}

\subsection{Related Work}
There has been considerable interest in developing clinical prediction models to estimate the risk of malignancy of suspicious breast lesions identified by ultrasound. Early contributions have focused on feature extraction and classical machine learning/statistical techniques \cite{Shen2007, Prabhakar2017AutomaticDA}, whereas recent work has concentrated around deep neural network-based computer vision approaches \cite{Kim2021, Qi2019, Shen2021, shen2023leveraging}.

Unfortunately, such attempts have often been plagued by at least one of the following issues: (i) lack of interpretability, (ii) lack of transparency/reproducibility, and (iii) lack of good software development principles. In particular, few systems go through the full software development and maintenance cycle and actually become accessible to doctors, and those that do are often based on black-box models such as deep convolutional nets \cite{deHond2022, rudin2019stop}. Moreover, most prediction models found on the literature are either poorly developed or poorly reported \cite{Steyerberg_2019, Collins2024}.

We stress that model interpretability is a crucial, non-negotiable requirement for prediction models in healthcare \cite{deHond2022, rudin2019stop, Steyerberg_2019}. While complex models such as deep neural nets may achieve slight improvements in predictive performance, these gains are often marginal when compared to the transparency offered by simpler models \cite{hastie2009elements, doshi2017towards}. Logistic regression is often highlighted as a reliable and interpretable model \cite{Steyerberg_2019}, where each coefficient can be directly interpreted as the contribution of a specific variable to the risk of cancer. This is significantly more direct than existing methods for interpreting neural networks \cite{rudin2019stop, lipton2018mythos}.

Interpretability is not only about understanding the model's mechanisms but also about quantifying the confidence in its predictions. Conformal prediction \cite{shafer2008tutorial, vovk2005algorithmic, angelopoulos2023conformal} is a method that provides valid measures of uncertainty for predictions in a variety of machine learning tasks, ensuring that the coverage probability of prediction intervals/sets is guaranteed under minimal assumptions.

In the field of healthcare, conformal prediction has shown significant promise \cite{survey_cp_health}. For instance, \citet{amnioml} employed conformal prediction to segment and predict the volume of amniotic fluid from fetal MRIs. Likewise, \citet{papadopoulos_health} utilized this method for diagnosing acute abdominal pain. Furthermore, \citet{nature_cp_health} applied conformal prediction for the diagnosis and grading of prostate biopsies.

Recent advances in conformal prediction have concentrated on improving local coverage properties, tailoring predictive intervals or sets to the specific characteristics of the data \cite{conformalized_quantile_reg, lei2012distributionfreepredictionbands,izbicki2020flexible,izbicki2022cd,cabezas2024regression}. These enhancements make distribution-free, model-agnostic uncertainty quantification tools more closely aligned with the precision medicine objectives that clinical prediction models seek to achieve \cite{Steyerberg_2019}.

\begin{figure}[t]
    \centering
    \begin{subfigure}[b]{0.4\textwidth}
        \centering
        \includegraphics[width=\textwidth]{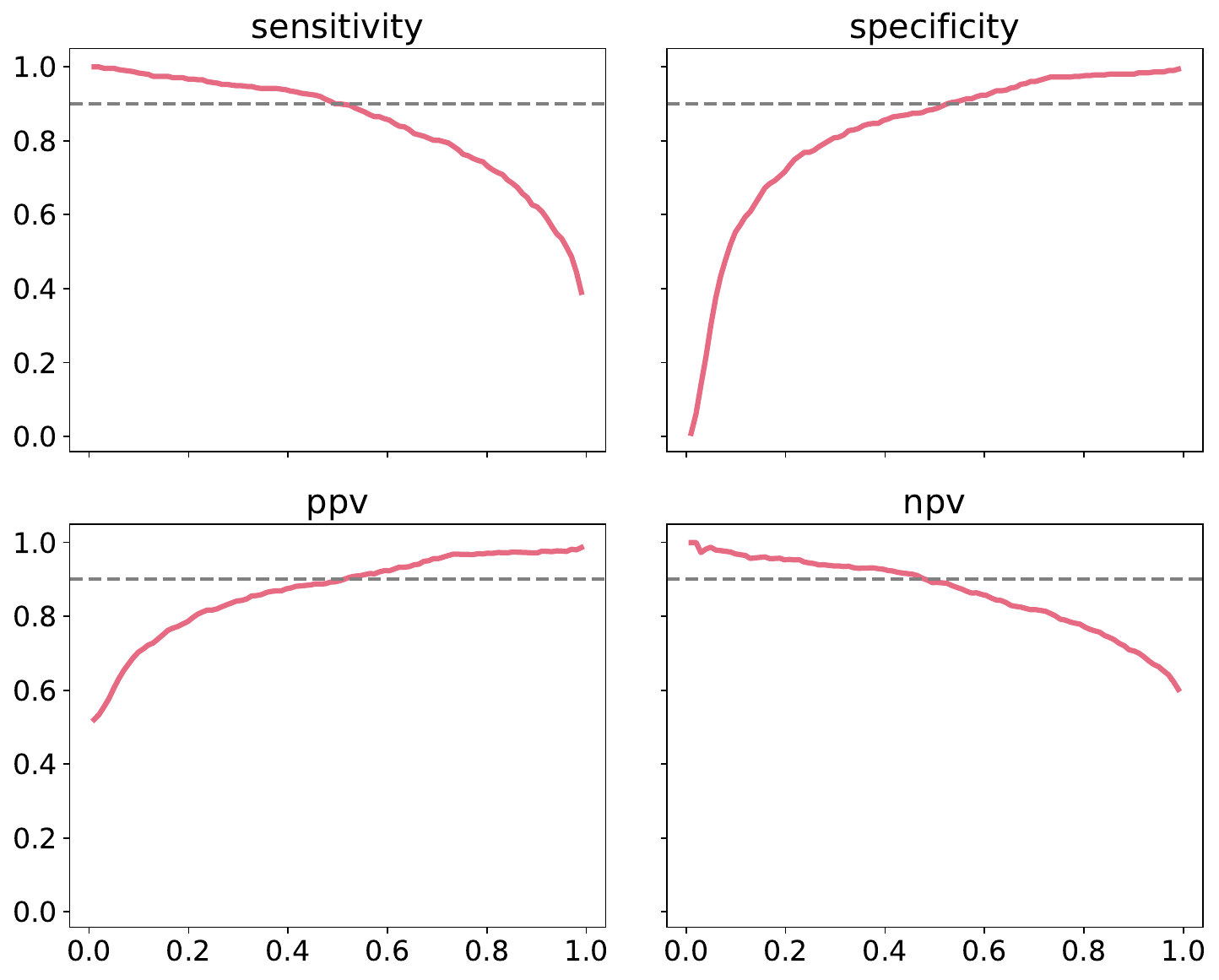}
        \caption{Performance metrics.}
        \label{fig:classification_curves}
    \end{subfigure}
    \hfill
    \begin{subfigure}[b]{0.4\textwidth}
        \centering
        \includegraphics[width=.8\textwidth]{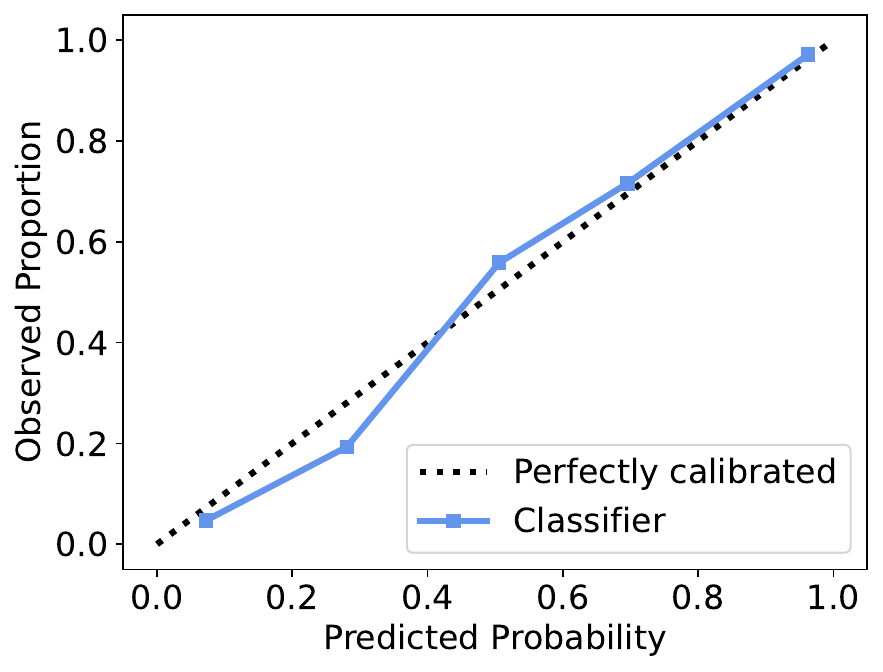}
        \caption{Calibration curve.}
        \label{fig:calibration_curve}
    \end{subfigure}
    \caption{Results of risk estimation on the calibration set: (a) Classification metrics and (b) Calibration curve.}
    \label{fig:lr_test_on_cal}
\end{figure}

\subsection{Notation}
Throughout the paper we will make use of the following notation. The set $\{(X_i, Y_i)\}_{i=1}^n$ will denote an independent and identically distributed (i.i.d.) sample of pairs of breast lesion data $X_i\subset \mathcal{X}\subset \mathbb{R}^D$ and their associated malignancy $Y_i\in\{0,1\}$ (benign or malignant, respectively). We define $[n] = \{1,\dots,n\}$. Data indices will be partitioned into $I_{\train} = [n_{\train}]$, $I_{\calib} = [n_{\train} + n_{\calib}] \setminus [n_\train] $, and $I_{\test} = [n]\setminus [n_\train + n_\calib]$, where $n_{\train} + n_{\calib} + n_{\test} = n$ and $n_{\train}, n_{\calib}, n_{\test} > 0$. Any model $\widehat{p}$ will always be trained on $\{(X_i, Y_i)\}_{i \in I_{\train}}$, have its output calibrated on $\{(X_i, Y_i)\}_{i \in I_{\calib}}$, and evaluated on $\{(X_i, Y_i)\}_{i \in I_{\test}}$.

\subsection{Dataset}
The Breast Lesion Dataset is comprised of $n = 1936$ breast lesions and contains both clinical, BI-RADS, and Doppler features.
The lesions come from a recent multicenter observational study with a retrospective and a prospective cohort of patients \cite{Buzatto2024}. The retrospective cohort included all patients from HCRP-USP (2014-2021) and MATER (2019-2021) who underwent diagnostic ultrasound followed by core needle or excisional biopsy, whereas the prospective cohort comprised similar patients from HCRP-USP, MATER, Hospital de Amor de Barretos, and Hospital de Amor de Campo Grande, starting in 2021 and is still ongoing.

The inclusion criteria were breast lesions classified as BI-RADS at least 3 identifed by ultrasound and submitted to percutaneous core needle biopsy and/or excisional biopsy. Exclusion criteria were lesions classifed as BI-RADS 2, lesion, size greater than 30mm, age less than 18 years old, pathologies not primary from the breast and lesions submitted exclusively to fine-needle aspiration cytology. The full details can be seen in \cite[Methodology]{Buzatto2024}.

In this paper, we introduced minor modifications to the dataset. Specifically, we used a slightly updated version of the dataset and excluded entries with missing data for the variables ``palpable'', ``age'', and ``ri'' instead of performing imputations. We also retained the original BI-RADS categories of ``shape'' and ``margins'' without collapsing them. The resulting dataset was then divided into training ($n_\train = 513$), calibration ($n_\calib = 1059$), and test ($n_\test = 364$) subsets. The training dataset consists solely of retrospective data, while the test dataset comprises only prospective data. The calibration dataset includes a mix of both, with its larger size being necessary to support the local conformal prediction method, which requires sufficient data per subgroup \cite{angelopoulos2023conformal}.

Prospective and retrospective patients have similar ages, with mean ages of $51.95 \pm 13.58$ and $51.44\pm 15.18$ years, respectively. Both groups have similar lesion size, with means of $16.70\pm 6.99$mm in the prospective and $16.74\pm 6.85$mm in retrospective. The proportion of palpable lesions is lower in the prospective group, at $55.12\%$, compared to $64.17$ in the retrospective. The resistance index (RI) is also slight lower in the prospective group, with a mean of $0.39\pm 0.38$, compared to $0.48\pm 0.41$ in the retrospective.
Irregular shapes are the most common in both groups, being more prevalent in the prospective group at $61.17\%$ compared to $51.80\%$. Circumscribed margins are more frequent in the retrospective data, representing $36.51\%$. Orientation is mostly parallel in both groups, with a slight higher proportion in the prospective group at $73.82\%$ compared to $67.19\%$ in the retrospective.

A table summarizing these statistics will be provided in the supplementary material. The resulting Breast Lesion Dataset will be available as a csv file.

\section{PersonalizedUS}
We introduce PersonalizedUS, an interpretable machine learning system for assessing the malignancy of breast lesions identified by ultrasound. The system combines regularized logistic regression with modern conformal prediction tools to deliver accurate (see Table~\ref{tab:baseline_results} and Figure~\ref{fig:lr_test_on_cal}) and interpretable predictions with personalized uncertainty quantification.

The system is deployed as an intuitive web application (see Figure~\ref{fig:deployed_system}) for specialist physicians at four Brazilian breast cancer reference centers: HCRP-USP, MATER, Hospital de Amor de Barretos, and Hospital de Amor de Campo Grande. The platform allows clinicians to input patient-specific data, receive interpretable risk assessments, and explore the detailed uncertainty measures in real-time. The interface is designed to be user-friendly, ensuring that even those with minimal technical expertise can effectively utilize the tool.

Two use cases of the system are described in ``Scope and Promise for Social Impact'', along with expected social benefits in terms of facilitated access to medical resources and reduced mental health burden among women with suspected breast cancer \cite{DRAGESET2003}.

A key innovation in our work is the adaptation of Locart \citep{cabezas2024regression}, a conformal prediction method designed for achieving local conditional coverage, to the binary classification setting of malignancy prediction. This novel approach involves fitting a decision tree regressor to a modified calibration dataset:

\[
\mathcal{D}_{\res} \coloneq \{(X_i, r_i) \mid i\in I_\calib\},
\]

\noindent where $r_i$ represent the classification residuals from the logistic model on the calibration set $\{(X_i, Y_i)\}_{i \in I_\calib}$. The resulting decision tree (see Figure~\ref{fig:dt}) partitions the lesion space into distinct subgroups, reflecting the model's varying difficulty in assessing these lesions. Surprisingly, our model captures well-known patterns that expert physicians recognize in US exams, such as circumscribed lesions in young patients, spiculated lesions with high vascularization, and other less-defined subgroups that are more challenging for doctors to manage.

Locart calibrates the risk assessment model locally within each partition using standard conformal prediction techniques \cite{angelopoulos2023conformal}. This approach generates personalized confidence sets for each prediction, giving clinicians both a risk estimate and a clear measure of uncertainty. As demonstrated by Theorems~\ref{thm:partition} and \ref{thm:consistency}, and shown empirically in Table~\ref{tab:cp_results}, the method achieves conditional coverage, unlike standard conformal prediction methods that only provide marginalized coverage.

The source code for PersonalizedUS, along with its documentation, will be published in a public repository after the anonymous review process. The web application will also be accessible following the review.

\section{Methodology}
\subsection{Risk Assessment Model}\label{sec:method_risk}
In our study, we trained a logistic regression model to assess the malignancy of breast lesions identified during ultrasound examinations. Logistic regression was chosen for its simplicity and interpretability, making it highly suitable for healthcare applications \cite{Steyerberg_2019}.

The model was developed on the training set following a grid search procedure based on 5-fold cross validation for optimizing the regularization parameter $C \in \{0.01, 0.1., ..., 100\}$. The model with the lowest log-loss was chosen as the best model.  The model was implemented in \texttt{scikit-learn}, with the the categorical features ``margins'' and ``shape'' being one-hot encoded and the numerical features ``age'', ``size'', and ``ri'' being standardized.

\subsection{Baselines}
To provide a robust evaluation of our logistic regression model, we compared its performance against several other widely-used machine learning algorithms, including decision trees, Naive Bayes, k-nearest neighbors (KNN), XGBoost, neural networks, AdaBoost, random forest, and support vector machines (SVM). Each model was trained and evaluated under the same experimental conditions to ensure a fair comparison.

The results are summarized in Table \ref{tab:baseline_results}. Our logistic regression model achieved a competitive performance, with a strong balance between sensitivity (0.900) and specificity (0.888). Although some more complex models showed slightly higher metrics, the differences were marginal. These results suggest that there are no significant advantages in using more complex models at the cost of the interpretability offered by logistic regression.

\begin{table}[h]
    \centering
    \setlength{\tabcolsep}{1mm}
    \begin{tabular}{lcccc}
        \toprule
        Model & Sensitivity & Specificity & PPV & NPV \\
        \midrule
        \textbf{Logistic} & 0.900 & 0.888 & 0.897 & 0.892 \\
        DT & 0.770 & 0.947 & 0.940 & 0.793 \\
        NB & 0.687 & \textbf{0.975} & \textbf{0.967} & 0.743 \\
        KNN & \textbf{0.903} & 0.894 & 0.902 & \textbf{0.896} \\
        XGB & 0.865 & 0.871 & 0.878 & 0.857 \\
        NN & 0.885 & 0.904 & 0.908 & 0.880 \\
        AB & 0.818 & 0.908 & 0.905 & 0.822 \\
        RF & 0.900 & 0.878 & 0.888 & 0.891 \\
        SVM & 0.898 & 0.896 & 0.903 & 0.891 \\
        \bottomrule
    \end{tabular}
    \caption{Comparison of learning algorithms over the Breast Lesion Dataset.}
    \label{tab:baseline_results}
\end{table}

\subsection{Personalized Uncertainty Quantification}

In this section, we present a novel method to construct interpretable and adaptive prediction sets for the predictions of our risk model. Concretely, we adapt the Locart method to the binary classification setting of malignancy prediction \cite{cabezas2024regression}.

The method is designed to partition the lesion space in a way that reflects the varying difficulty of malignancy prediction with respect to our machine learning model. Standard conformal prediction methods are then applied to these local partitions, thereby providing local coverage guarantees for individual predictions. The details of this approach are outlined next.

\begin{enumerate}
    \item \textbf{Calculating residuals.} Let $\widehat{p}$ be the malignancy prediction model. For each $i\in I_{\calib}$, we began by calculating the residuals over the calibration set:
    \[
    r_i = \begin{cases}
        1 - \hat{p}(Y=1|X_i),\quad \textrm{if}\quad y_i = 1, \\
        1 - \hat{p}(Y=0|X_i),\quad \textrm{if}\quad y_i = 0.
    \end{cases}
    \]

    \item \textbf{Creating residual dataset.} A new dataset \( \mathcal{D}_{res} \) was then created by combining the calibration data \( \mathcal{D}_{cal} \) with the residuals \( r_i \), that is:
    \[
    \mathcal{D}_{\res} \coloneq \{(X_i, r_i) \mid i\in I_\calib\}.
    \]
    Then, we split this dataset in two subsets: $\mathcal{D}_{\res, 0}$ and $\mathcal{D}_{\res, 1}$, such that, $\mathcal{D}_{\res}=\mathcal{D}_{\res,0}\cup \mathcal{D}_{\res, 1}.$

    \item \textbf{Learning patient subgroups.} We trained a decision tree regressor $\widehat{T}$ over $\mathcal{D}_{\res,0}$.
    This decision tree induced a partition of the feature space with size $K$:

    \[\mathcal{X}=\bigcup_{i=1}^K\mathcal{X}_i.\]

    We expect this partition to effectively capture and reflect the varying complexity of cancer predictions within each lesion subgroup, ensuring that the risk model's local behavior is accurately represented.

    \item \textbf{Calculating quantiles for each subgroup.} Given a miscoverage level $\alpha \in (0,1)$, for each partition learned from $\mathcal{D}_{\res,0}$, we calculated an adjusted coverage quantile $\tilde{\alpha}$ over $\mathcal{D}_{\res, 1}$. Specifically, for each element $\mathcal{X}_j$ of the partition, the quantile \( q_j \) was determined as:
    \[
    q_{j,1-\alpha} = \textrm{Quantile}_{1-\tilde{\alpha}}\left( r_i :\,  (X_i,r_i)\in \mathcal{D}_{\res,1},\  X_i\in \mathcal{X}_j \right),
    \]
    where
    \[ \tilde{\alpha} = \frac{\lceil(k_j + 1)\cdot\alpha\rceil}{|k_j|}, \]

    is the adjusted significance level of $\alpha\in(0,1)$ \cite{angelopoulos2023conformal}     and
    \[
    k_j \coloneqq \left|\{(X_i, r_i) \in \mathcal{D}_{\text{res},1} \mid X_i \in \mathcal{X}_j\}\right|
    \]
is the number of patients in $\mathcal{D}_{\res,1}$ that are also in $\mathcal{X}_j.$

    \item \textbf{Personalized quantification of uncertainty.} Given a test lesion $X_i,\ i\in I_\test$, we check which element $\mathcal{X}_j$ of the partition it belongs, then we define its prediction set as
    \[C_{\locart}(x)=\{\ell\in\{0,1\}: \widehat{p}(y=\ell|X=x)\geq 1-q_{j,1-\alpha}\}.\]

\end{enumerate}

The theoretical foundations for this method are established in \cite{cabezas2024regression} and explained next.

\begin{theorem}[Theorem 2 in \citet{cabezas2024regression}]\label{thm:partition}
Let \(\{\mathcal{X}_1, \ldots, \mathcal{X}_K\}\) be a finite partition of the lesion space. Given an exchangeable sequence \((X_i, y_i)_{i=1}^{n+1}\) and a miscoverage level \(\alpha \in (0,1)\), the following holds:
\[
\mathbb{P}\left[Y_{n+1} \in C_{\locart}(X_{n+1}) \mid X_{n+1} \in \mathcal{X}_j\right] \geq 1 - \widehat{\alpha},
\]
for all \(j = 1, \ldots, K\) where $ \tilde{\alpha} = \frac{\lceil(k_j + 1)\cdot\alpha\rceil}{|k_j|}$
and     $k_j \coloneqq \left|\{i \in [n] \mid X_i \in \mathcal{X}_j\}\right|$.
\end{theorem}

The theorem guarantees that, given a finite partition of the feature space, the probability that the prediction for a new sample falls within the local prediction band—conditioned on the sample belonging to one of these partitions—is at least \(1 - \alpha\).

The core idea behind the in this approach model is that, by leveraging the partition learned from the decision tree, we can achieve local conditional coverage. In fact, \citet{cabezas2024regression} show that this local method for uncertainty quantification achieves asymptotic coverage.

\begin{theorem}\label{thm:consistency}
Given a lesion $x\in\mathcal{X}$, let $\mathcal{X}(x)$ be the element of the learned partition given by Locart which $x$ belongs to. Under certain regularity conditions, Locart has conditional validity, that is
\begin{align*}
    \lim_{n\to \infty}\mathbb{P}\left[Y_{n+1} \in C_\locart(X_{n+1}) \mid X_{n+1} \in \mathcal{X}(x)\right]=\\
    \mathbb{P}\left[Y_{n+1} \in C(X_{n+1}) \mid X_{n+1} =x\right],
\end{align*}
where $C(\cdot)$ is the theoretical optimal confidence set.
\end{theorem}

In summary, integrating Locart into PersonalizedUS marks a significant step forward in uncertainty quantification for breast cancer risk assessment. This approach ensures that each prediction comes with a locally adaptive confidence interval, enhancing both the reliability and interpretability of the model, ultimately making it a more effective tool for clinical decision-making.

\subsection{Learning Lesion Subgroups}

Following the steps outlined in the previous section, we fitted a decision tree regressor to the residual dataset. The tree model was developed on the basis of a grid search procedure via 5-fold cross-validation for optimizing the structural parameters $\text{``max\_depth''} \in [3, 4, 5, 6]$ and $\text{``min\_samples\_leaf''} \in [70, 80, 90, 100]$. The model with the lowest mean squared error was chosen as the best model. The model was implemented in \texttt{scikit-learn}, with the categorical features ``shape'' and ``margins'' being ordinal-encoded.

\section{Evaluation}
\subsection{Risk Estimation}\label{sec:eval_risk}

The selected logistic regression model was then evaluated on the calibration set to determine its performance. We compared the model's risk estimates to the actual outcomes and calculated key performance metrics, including sensitivity (recall of class 1), specificity (recall of class 0), positive predictive value (precision of class 1), and negative predictive value (precision of class 0), the resulting curves can be seen in Figure~\ref{fig:lr_test_on_cal} along with the model's calibration curve. We also evaluated the Area Under the ROC curve  (0.958), the Area Under the Precision-Recall Curve (0.960), the and log-loss (0.268).

\subsection{Lesion Subgroup Evaluation}

\begin{figure*}[t]
    \centering
    \includegraphics[width=0.75\textwidth]{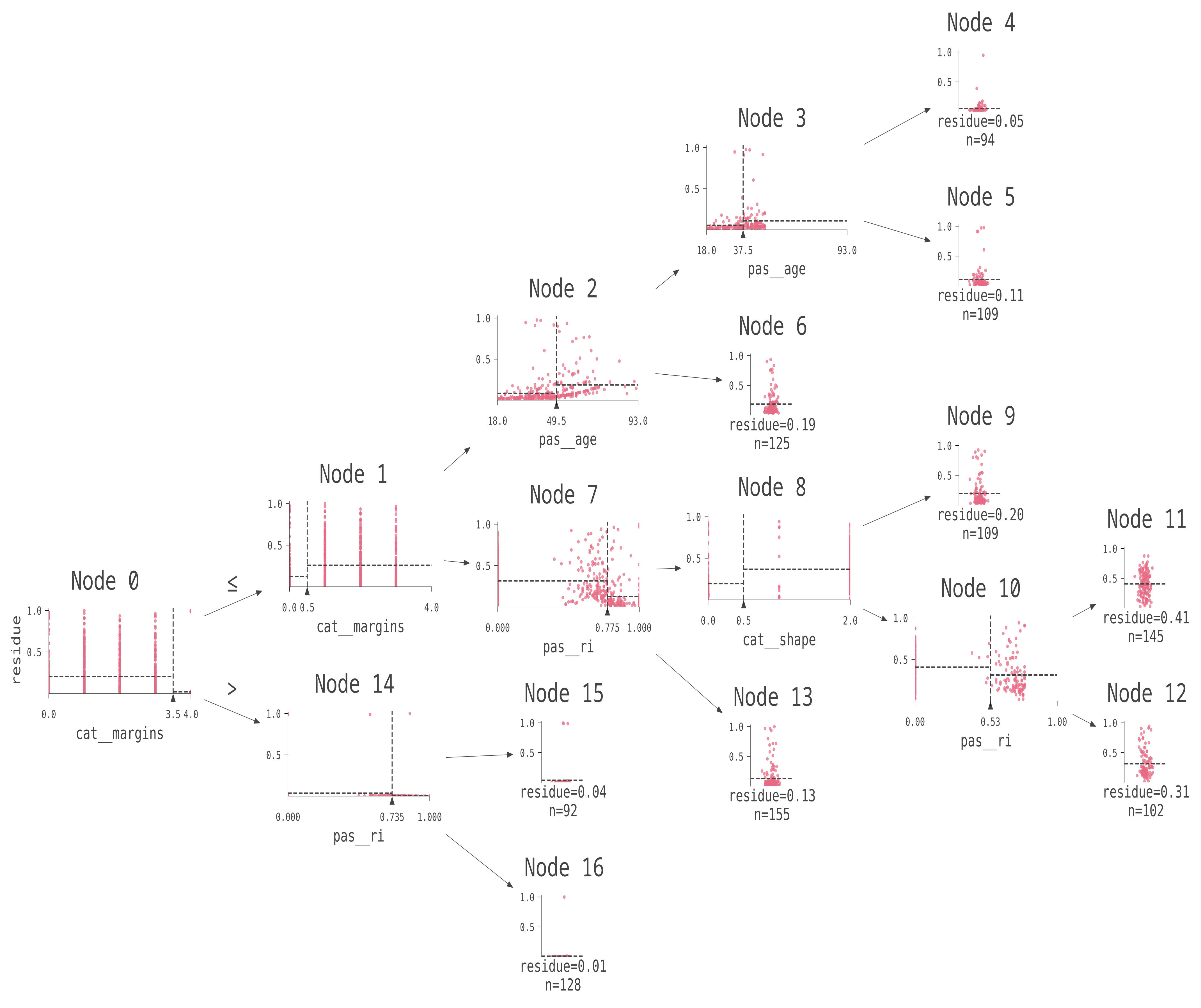}
    \caption{Decision tree regressor trained for predicting the classification residuals of our logistic regression model over the calibration set.}
    \label{fig:dt}
\end{figure*}

Two medical doctors with specialized training in AI (IB and DT) evaluated the resulting tree structure depicted in Figure~\ref{fig:dt}. According to their expert assessment, the model successfully captures patterns that physicians commonly recognize in ultrasound exams, such as circumscribed lesions in young patients, spiculated lesions with high vascularization, and other less-defined subgroups that are more challenging to manage.

\begin{figure}[ht!]
    \centering
    \includegraphics[width=0.8\columnwidth]{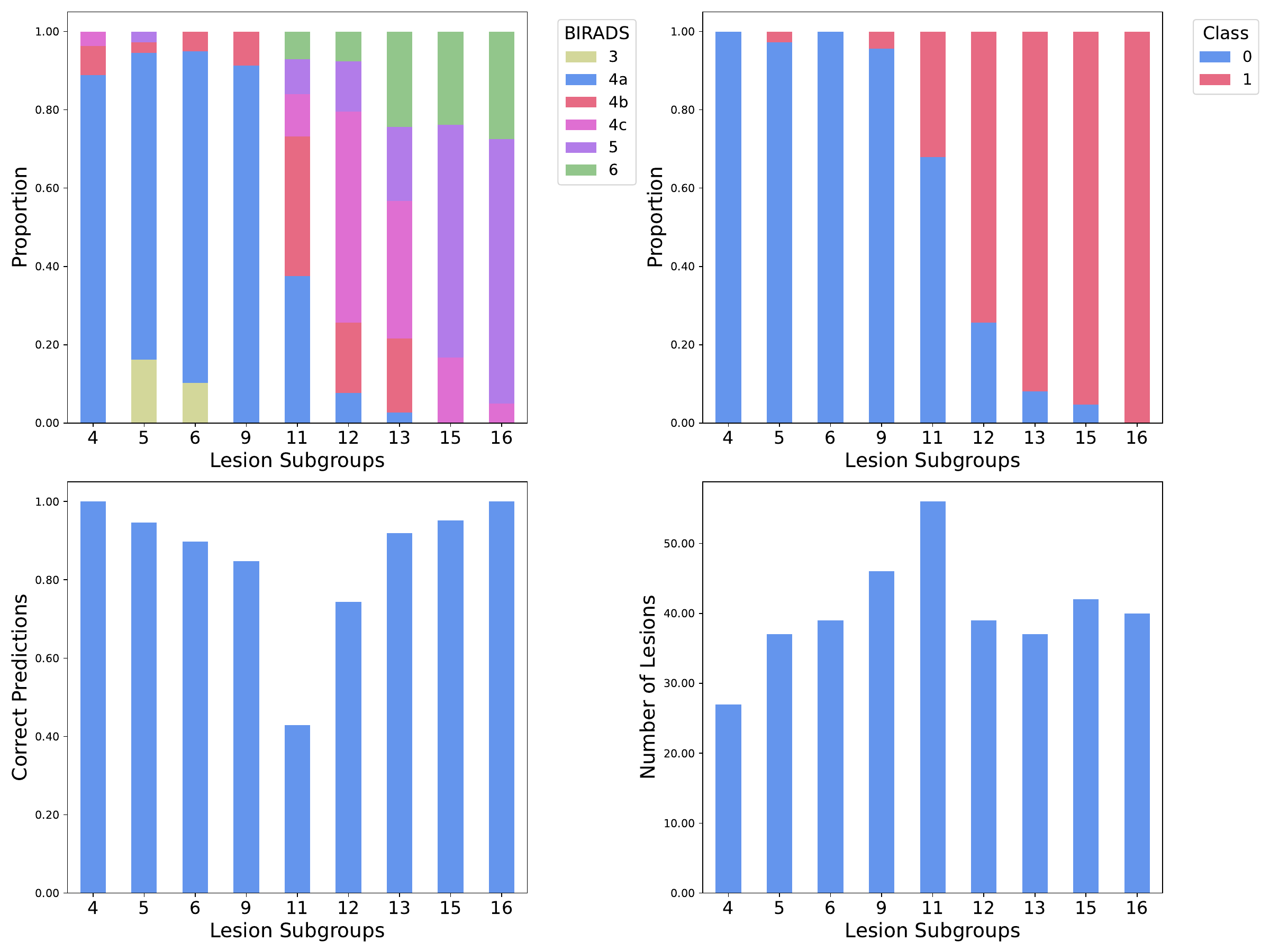}
    \caption{Lesion subgroups analysis of BIRADS categories, malignancy, predictive accuracy, and subgroup size.}
    \label{fig:combined_leaves}
\end{figure}

As illustrated in Figure~\ref{fig:combined_leaves}, these patterns are also evident in the distribution of BI-RADS categories and malignancy across subgroups. The leftmost groups predominantly feature benign BI-RADS 3 and 4a lesions, while the rightmost groups contain most BI-RADS 4c, 5, and malignant lesions, which are generally easier to predict. The middle groups, however, are more mixed, consisting mainly of BI-RADS 4 and 5 lesions with varying levels of malignancy. These lesions proved especially challenging for our logistic regression model.

\subsection{Uncertainty Quantification Evaluation}

\begin{figure}[th]
    \centering
    \includegraphics[width=.8\columnwidth]{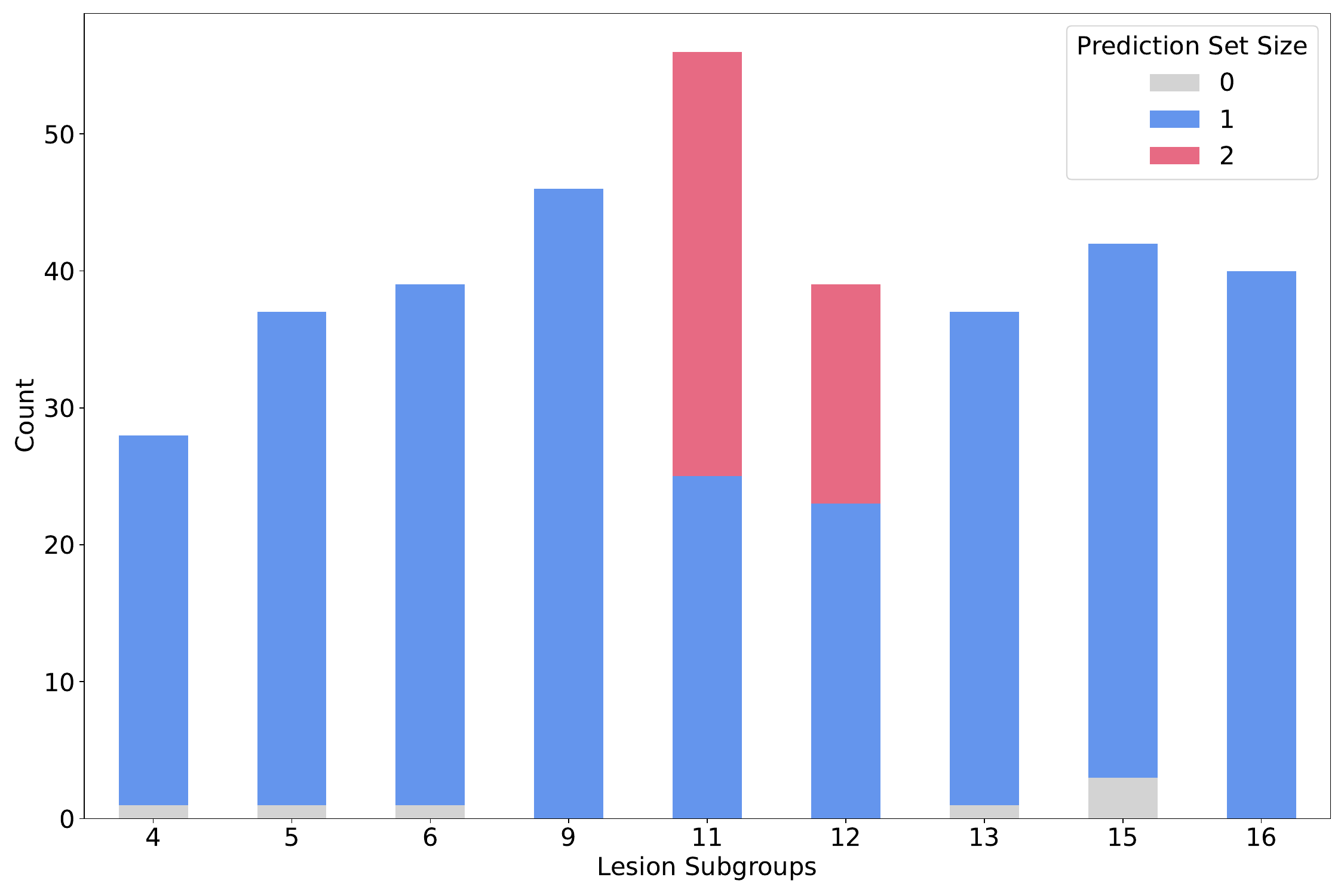}
    \caption{Stacked plot of (local) average set sizes.}
    \label{fig:stacked_sum_thresholds}
\end{figure}

Prediction sets generated by Locart were evaluated according to three criteria: (i) average set size, (ii) comparison between theoretical and empirical coverage, and (iii) proportion of prediction sets containing only the ground truth. A successful application of conformal prediction to binary classification is expected to often produce singleton sets, with empirical coverage close to $1-\alpha$. A successful application of Locart further requires that these metrics hold conditionally to $X_j\in \mathcal{X}_J$ for $j\in I_\test$. As such, in Table~\ref{tab:cp_results} we show the result of applying the three evaluation criteria with $\alpha=0.1$, grouping by leaf. Also, we show a stacked plot of set sizes in Figure~\ref{fig:stacked_sum_thresholds}.

As we can see from the first column of Table~\ref{tab:cp_results} and Figure~\ref{fig:stacked_sum_thresholds}, the average set size was close to 1 for most leaves, except for leaves 11 and 12. This pattern is also evident in the last column of Table~\ref{tab:cp_results}, where lesions within leaves 11 and 12 were considerably more challenging for our risk model to assess. This behavior is expected, as Locart groups lesions based on their relative difficulty for the malignancy prediction model. Therefore, leaves 11 and 12 contain the more challenging lesions, enabling tighter predictive sets in the remaining leaves. Notably, these two ``hard'' leaves represent only 20\% of the test set, suggesting that our model reliably predicts malignancy for 80\% of the dataset, which, like the entire dataset, consists solely of suspicious lesions that were biopsied. Finally, the observed fluctuation of up to 10\% in empirical coverage falls within the expected range, given the sample size of approximately 100 points per leaf \cite{angelopoulos2023conformal}.

\begin{table}[th]
\centering
\setlength{\tabcolsep}{1mm}
\begin{tabular}{lrrr}
\toprule
 & Avg. Set Size & Emp. Cov. (\%) & Truth Only (\%) \\
Leaf &  &  &  \\
\midrule
4 & 0.96 & 96.43 & 96.43 \\
5 & 0.97 & 94.59 & 94.59 \\
6 & 0.97 & 97.44 & 97.44 \\
9 & 1.00 & 95.65 & 95.65 \\
11 & 1.55 & 89.29 & 33.93 \\
12 & 1.41 & 94.87 & 53.85 \\
13 & 0.97 & 89.19 & 89.19 \\
15 & 0.93 & 90.48 & 90.48 \\
16 & 1.00 & 100.00 & 100.00 \\
\bottomrule
\end{tabular}
\caption{Evaluation of our local conformal prediction tool.}
\label{tab:cp_results}
\end{table}

\section{Scope and Promise for Social Impact}

Two significant social issues arise from false positive US evaluations: (i) an increased burden on the healthcare system, particularly on pathologists, which delays timely access to treatment for those in need, and (ii) the substantial mental health impact on patients and their families, especially in underserved communities \cite{DRAGESET2003}. This section discusses two potential applications of our prediction model aimed at optimizing the biopsy referral process to ensure it prioritizes those who really need it, and alleviate the stress and anxiety experienced by patients and their families.

The first, most straightforward application of our model is to assist radiologists in differentiating between benign and malignant lesions. In this scenario, the physician would initially evaluate the US exam and then, specially for lesions classified as BI-RADS 4a and 4b, use our machine learning system for a second opinion, if desired. In cases where the model is highly confident the lesion is benign, the radiologist would be prompted to carefully reconsider whether a biopsy referral is necessary.

Another, less intrusive application of our prediction model is to rank patients awaiting biopsy based on their assessed risks as determined by our model. Even though this approach wouldn't reduce the overall load on the healthcare system, it would help ensure that high-risk patients receive priority access to medical resources.

To evaluate the effectiveness of our system in differentiating between benign and malignant breast masses, we further analysed its test set performance by considering only BI-RADS 4a and 4b lesions. Using an approach similar to that of \citet{Buzatto2024}, we selected a threshold that maximized the negative predictive value while ensuring the positive predictive value did not fall below that of BI-RADS 4b \cite{Yoon2011}. If the model's recommendations were followed, only 64 out of 196 biopsies would be requested, representing a reduction of more than 65\%. Moreover, with the optimized threshold ($t = 0.223$), the model wouldn't miss any cancer cases among BI-RADS 4a and 4b lesions.

\section{Conclusion and Follow-up Work}

This paper introduces PersonalizedUS, an interpretable AI system for predicting malignancy of breast lesions identified by US. The system combines logistic regression with new conformal prediction methods to provide accurate predictions with individualized uncertainty quantification.

The system is deployed as a web application for specialists at four Brazilian breast cancer reference centers. The site allows doctors to enter patient data, receive interpretable risk estimates, and explore uncertainty quantification. We explored two use cases of the system, along with expected social benefits in terms of facilitated access to medical resources and reduced mental health burden among women with suspected breast cancer.

A novelty in our work is the use of Locart, a conformal prediction method designed for achieving local conditional coverage.
This method partitions the lesion space, reflecting the model's varying difficulty in assessing these lesions. Our tree captures well-known patterns that specialists recognize in US exams, such as circumscribed lesions in young patients, spiculated lesions with high vascularization, and other less-defined subgroups that are harder for doctors to manage.

\subsubsection{Facilitation of Follow-up Work}
Locart's model-agnostic nature allows our logistic regression model to be easily replaced by more complex models, such as neural networks, without affecting the personalized uncertainty quantification. Also, we will make both our data and code publicly available after the review process.



\bibliography{ourbib}
\end{document}